\documentclass[letterpaper]{article} 
\usepackage{aaai23}  
\usepackage{times}  
\usepackage{helvet}  
\usepackage{courier}  
\usepackage[hyphens]{url}  
\usepackage{graphicx} 
\usepackage{natbib}  
\usepackage{caption} 
\usepackage{algorithm}
\usepackage{algorithmic}

\usepackage{newfloat}
\usepackage{listings}
\DeclareCaptionStyle{ruled}{labelfont=normalfont,labelsep=colon,strut=off} 
\floatstyle{ruled}
\newfloat{listing}{tb}{lst}{}
\floatname{listing}{Listing}
\title{SolderNet: Towards Trustworthy Visual Inspection of Solder Joints in Electronics Manufacturing Using Explainable Artificial Intelligence}
\author{
    Hayden Gunraj\textsuperscript{\rm 1,2},
    Paul Guerrier\textsuperscript{\rm 3},
    Sheldon Fernandez\textsuperscript{\rm 2}, and
    Alexander Wong\textsuperscript{\rm 1,2}
}
\affiliations{
    \textsuperscript{\rm 1}Vision and Image Processing Research Group, University of Waterloo, Waterloo, Canada\\

    \textsuperscript{\rm 2}DarwinAI Corp., Waterloo, Canada\\

    \textsuperscript{\rm 3}Moog Inc., New York, USA\\
    \{hayden.gunraj, a28wong\}@uwaterloo.ca
}

\begin{document}

\maketitle

\begin{abstract}
    In electronics manufacturing, solder joint defects are a common problem affecting a variety of printed circuit board components. To identify and correct solder joint defects, the solder joints on a circuit board are typically inspected manually by trained human inspectors, which is a very time-consuming and error-prone process. To improve both inspection efficiency and accuracy, in this work we describe an explainable deep learning-based visual quality inspection system tailored for visual inspection of solder joints in electronics manufacturing environments. At the core of this system is an explainable solder joint defect identification system called \textbf{SolderNet} which we design and implement with trust and transparency in mind. While several challenges remain before the full system can be developed and deployed, this study presents important progress towards trustworthy visual inspection of solder joints in electronics manufacturing.\footnote{This paper consists of general capabilities information that is not defined as controlled technical data under ITAR Part 120.10 or EAR Part 772.}
\end{abstract}

\section{Introduction}

\begin{figure}[t]
    \centering
    \includegraphics[width=1\columnwidth]{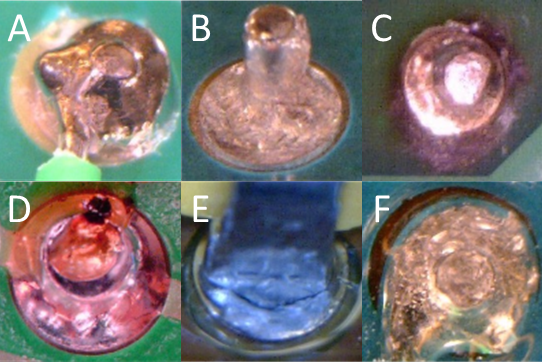}
    \caption{Example images of solder joint defects from the dataset examined in this study: (A) fractured joint, (B) cold joint, (C) burns, (D) flux residue, (E) poor wetting, and (F) disturbed solder.}
    \label{defects}
\end{figure}

In electronics manufacturing, solder joint defects are a common problem affecting both surface-mounted and through-hole components of printed circuit boards (PCBs) (see Figure~\ref{defects} for example solder joint defects). Defects introduced in the soldering process can lead to electrical issues and faulty parts, especially if not caught early in the process. This is particularly concerning in critical applications such as the aerospace and medical industries, where defective PCBs can cause catastrophic failures in critical systems. If defects are caught early, the solder can be reworked to minimize potential issues later in the manufacturing process and avoid unnecessary electronic waste.

To identify and correct solder joint defects, the solder joints on a PCB are often visually inspected by trained inspectors. However, human inspectors are estimated to make visual inspection errors in 20-30\% of cases, and the performance of human inspectors varies considerably depending on experience, mental fatigue, defect type, frequency of defect occurrence, and a variety of other factors~\cite{inspection1,inspection2}. Manual inspection is also expensive and time-consuming, often requiring multiple inspectors to achieve reasonable throughput.

\begin{figure*}[t]
    \centering
    \includegraphics[width=0.8\textwidth]{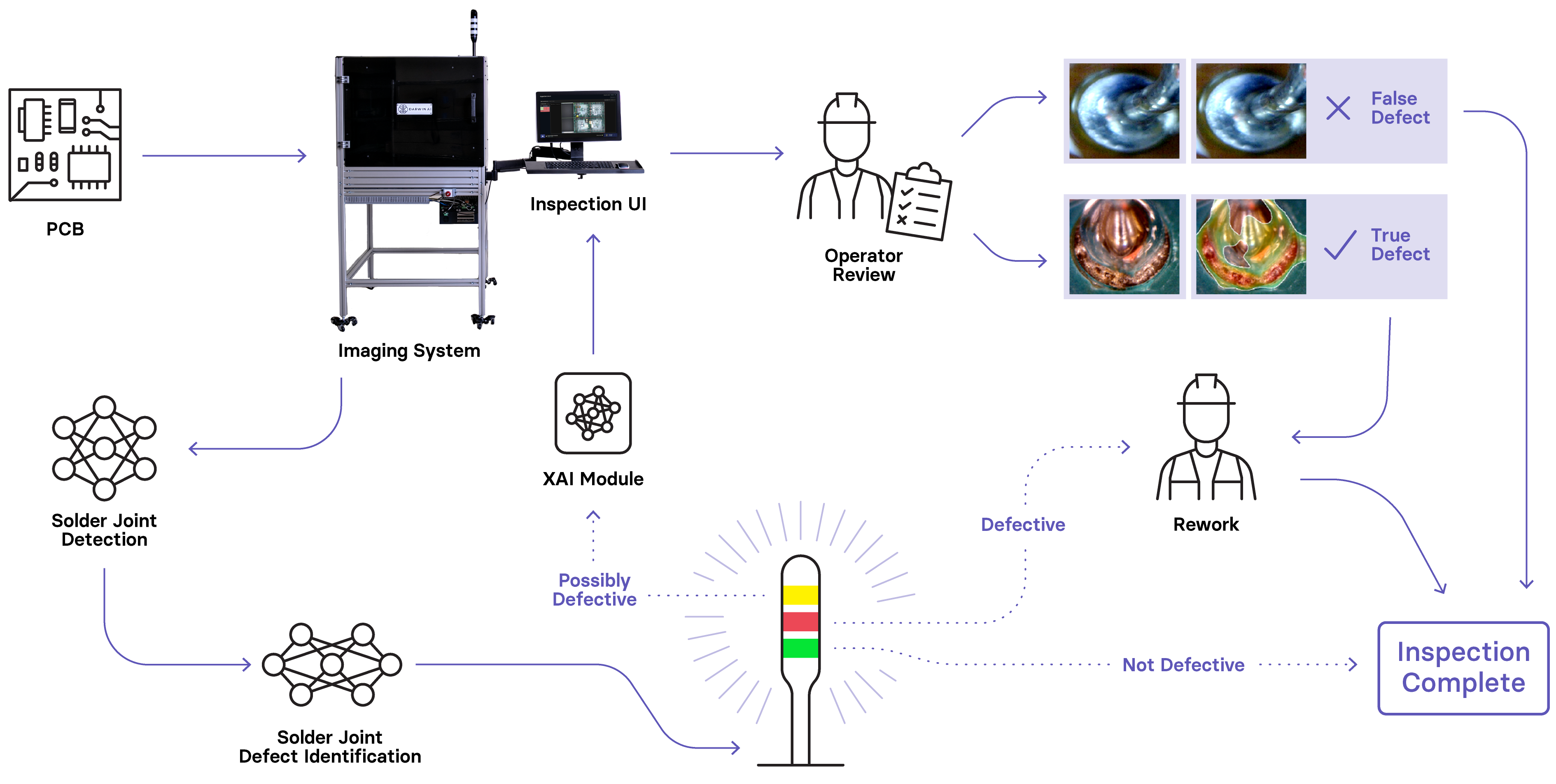}
    \caption{Overview of the proposed solder joint inspection system. This study focuses on developing the core defect identification system (SolderNet), which comprises the defect identification and XAI modules.}
    \label{fig:system}
\end{figure*}
In contrast to manual inspection, automated visual inspection of solder joints offers many benefits, such as high throughput, high performance, and zero fatigue. However, automating this task is technically challenging due to the small size of the joints, the wide variety of possible joint and defect types, limited computing resources, limited inspection time, and the need for high performance. Deep neural networks offer an attractive solution to this problem, as they have been successfully applied to detection and inspection tasks in a variety of manufacturing scenarios~\cite{autoinsp1,autoinsp2,autoinsp3,autoinsp4,autoinsp5}. While deep neural networks are capable of achieving human-level performance in visual inspection tasks, they have several drawbacks in critical manufacturing scenarios:
\begin{enumerate}
    \item \textbf{Computational complexity}: high-performance neural networks are often compute-hungry and require specialized hardware for fast inference. When computing resources are limited (as in many manufacturing scenarios), such networks can be slow and reduce throughput.
    \item \textbf{Lack of explanations}: neural networks typically operate as black boxes which provide a prediction but cannot explain how the prediction was made. In critical inspection scenarios, determining why a network made a particular prediction falls to the human inspector, which reduces the potential throughput benefits offered by these networks.
    \item \textbf{Unclear reliability}: the development of neural networks usually involves testing using data not seen by the network. Despite this, a common challenge when deploying deep neural networks is understanding when a model's predictions can be trusted and which scenarios a model is likely to fail in.
\end{enumerate}

In this work, we outline a trustworthy, explainable deep learning-driven solder joint inspection system for electronics manufacturing. The proposed system is capable of handling both through-hole and surface-mounted components and can provide explanations of its decision in order to facilitate manual review when necessary. We perform a number of experiments to implement and test the core functionality of this system with a focus on practical considerations for manufacturing settings. Moreover, we leverage a mix of quantitative explainability and trust quantification techniques to further analyze the behaviour and trustworthiness of the system, identify gaps in the model development process, and provide insights during the inspection process.

\section{Methods}

The proposed system consists of several stages which are designed to provide high performance, high throughput and high reliability. Figure~\ref{fig:system} illustrates the inspection system, which proceeds via the following steps:
\begin{enumerate}
    \item Images of a PCB are captured by an imaging system.
    \item PCB images are passed through a solder joint detector which identifies and extracts solder joint images.
    \item Solder joint images are passed through a defect identification network which categorizes solder joints as defective, possibly defective, or non-defective and activates the appropriate indicator.
    \item For boards with possibly defective joints, an XAI algorithm is used to generate explanations of the network's predictions.
    \item A human operator reviews the possibly defective joints and their XAI visualizations to determine which are truly defective and which are not defective.
    \item All defective solder joints are reworked.
\end{enumerate}

In this study, we focus on defect identification, explainability, and trustworthiness. These three aspects of the system are described in detail in the following subsections.

\subsection{Solder Joint Defect Identification}

We focus on convolutional neural networks (CNNs) to perform the task of solder joint defect identification. CNNs are a ubiquitous technology in image identification tasks, and recent advances in network design have enabled the creation of high-performance networks which remain computationally efficient. We examined both efficiency-focused architectures and performance-focused architectures to examine their respective trade-offs between network complexity and network performance. Specifically, we tested Attend-NeXt (Small and Large)~\cite{attendnext}, MobileNet (V2 and V3 Small)\cite{mobilenetv2, mobilenetv3}, ShuffleNetV2~\cite{shufflenetv2}, and ConvNeXt (Tiny, Small, and Base)~\cite{convnext}. Table~\ref{tab:arch} shows the number of parameters and floating-point operations (FLOPs) for each of the network architectures examined in this work. We train each of the aforementioned architectures as binary classifiers which provide a confidence score on the interval [0, 1] indicating the network's confidence that an input image represents a defective solder joint. In deployment, this allows us to define confidence regions which correspond to definite defects (requiring repair), possible defects (requiring review), and definite non-defects (no action required).


\subsection{Quantitative Explainability}

Following defect identification, as can been seen in Figure~\ref{fig:system} some solder joints may require review by a human operator given a level of uncertainty during the solder joint defect identification process with respect to whether that joint is indeed defective. In such scenarios, while the identification network's prediction and confidence provide useful information to the operator, the operator must still identify the particular defects (or lack thereof) in the solder joint images to determine if it is a false defect detection (in which case inspection is complete) or if it is indeed a true defect detection (in which case the solder needs to be reworked). To make this task easier and faster for operators, we propose a quantitative explainability module which identifies the critical factors in an image which led to the neural network's decision in a quantitative manner, allowing the operator to rapidly identify the locations of potential defects and validate the model's predictions.

To this end, we introduce an extended form of GSInquire~\cite{gsinquire} to provide visual explanations of a neural network's decision-making process. We choose GSInquire as the core approach to extend upon because it identifies specific critical factors that quantitatively impact the decisions made by the deep neural network, in contrast to other explainability methods such as Grad-CAM~\cite{gradcam}, Expected Gradients~\cite{expgrad}, LIME~\cite{lime}, and SHAP~\cite{shap} that generate only qualitative heatmaps depicting relative importance.

In its original form, GSInquire examines a network's activation signals in response to an input image and uses them to identify critical factors within the image which impact the network's decision in a quantitatively significant way. These critical factors may then be projected into the same space as the image to produce a visual interpretation. Building upon GSInquire, we introduce an extension which determines the relative importance of different aspects within the critical factors of a given image. This allows for more fine-grained interpretation of the critical factors, as we can now see not only which critical factors contribute the most to the neural network's decision-making process but also which aspects of a particular critical factor are most important. We present the relative importance of different aspects within the critical factors as regional heatmap overlays within the boundaries of their corresponding critical factors, as shown in Figure~\ref{fig:xai}.

\begin{table}[t]
    \centering
    \begin{tabular}{lll}
        Architecture & Parameters (M) & FLOPs (G)\\
        \hline\\[-1.5ex]
        Attend-NeXt Small & 1.632 & 0.423\\
        Attend-NeXt Large & 3.897 & 0.861\\
        MobileNetV2 & 2.225 & 0.409\\
        MobileNetV3 Small & 1.519 & 0.076\\
        ShuffleNetV2 & 1.255 & 0.193\\
        ConvNeXt Tiny & 27.821 & 5.832\\
        ConvNeXt Small & 49.455 & 11.364\\
        ConvNeXt Base & 87.567 & 20.084        
    \end{tabular}
     \caption{Comparison of architectural and computational complexity for each of the tested network architectures. FLOPs were measured with a 256$\times$256 3-channel input and a batch size of 1.}
     \label{tab:arch}
\end{table}

\begin{table*}[t]
    \centering
    \begin{tabular}{llllll}
        Architecture & Latency (s) & Accuracy (\%) & Overkill (\%) & Escape (\%) & NetTrustScore\\
        \hline\\[-1.5ex]
        Attend-NeXt Small & \textbf{0.275} & 86.6 & 8.4 & 5.0 & 0.864\\
        Attend-NeXt Large & 0.430 & \textbf{91.1} & \textbf{5.0} & 3.9 & \textbf{0.904}\\
        MobileNetV2 & 4.082 & 88.5 & 7.8 & \textbf{3.7} & 0.878\\
        MobileNetV3 Small & 1.647 & 88.3 & 7.1 & 4.6 & 0.883\\
        ShuffleNetV2 & 1.775 & 87.0 & 7.1 & 5.9 & 0.867\\
        ConvNeXt Tiny & 5.975 & 88.8 & 7.2 & 3.9 & 0.887\\
        ConvNeXt Small & 12.027 & 89.6 & 5.8 & 4.6 & 0.890\\
        ConvNeXt Base & 17.575 & 88.8 & 6.3 & 4.8 & 0.889
    \end{tabular}
     \caption{Comparison of quantitative performance metrics for each of the network architectures on the solder joint test dataset. Latency was measured on an ARM Cortex-A72 with a 256$\times$256 3-channel input and a batch size of 1, and is reported as the average of 100 measurements taken after 20 warm-up runs. Accuracy, overkill, and escape were calculated at a confidence threshold of 0.5. Best results highlighted in \textbf{bold}.}
     \label{table:quant}
     \vspace{-0.07in}
\end{table*}

\subsection{Second-order Explainability}

In addition to providing visual explanations in deployment settings, visual explainability is also a valuable model validation tool during development. While quantitative metrics such as accuracy provide important measures of a deep neural network's performance, they do not provide information regarding how decisions are made. To address this gap in performance analysis, visual XAI enables auditing of a model's decisions during development to ensure that they are based on relevant visual indicators and elucidate potential biases in the training data, which may then be used to guide improvements to the training framework. However, reviewing visual explanations manually is a time-consuming task, particularly for large-scale image datasets with many classes or high intra-class variability. Manual review may also be influenced by human biases, and it can be challenging if not intractable to mentally conceptualize key trends and patterns based on the individual explanations at hand.

To facilitate auditing of the model and dataset during development, we introduce the concept of second-order explainable artificial intelligence (SOXAI) which extends the concept of XAI from the sample level to the dataset level. Rather than manually reviewing visual explanations to explore patterns in a model's decision-making behaviour, SOXAI aims to reveal these patterns automatically through analysis of the relationships between quantitative explanations. This allows for rapid identification of the most common visual concepts leveraged by a model when making its decisions, and can reveal obvious model and dataset biases. This can also increase transparency by helping users understand what a model has learned and what it has not. In essence, SOXAI enables us to ''explain explainability'' by providing higher-level interpretations of the behaviours of deep neural networks.

In this study, we formulate SOXAI as an embedding problem: given an image $I$ and corresponding quantitative explanation $\alpha$, we define an embedding $f:(I, \alpha)\rightarrow \mathbf{R}^N$ which embeds the regions of $I$ indicated by $\alpha$. Performing this embedding for all images in a dataset allows for similar embeddings to be grouped together with secondary algorithms. To generate the visualizations presented in this work, we leverage t-distributed stochastic neighbour embedding (t-SNE)~\cite{tsne} to group the embeddings and map them to a 2-dimensional space.

\subsection{Trust Quantification}

In industrial applications, it is important to understand how trustworthy a model is in order for it to be deployed as an automation tool. Quantifying trust allows for models to be compared in terms of their trustworthiness during the model development process, and may help guide decisions as to the level of human supervision and review required once a model is deployed. Recent studies by Wong et al.~\cite{trust1,trust3} and Hyrniowski et al.~\cite{trust2} introduced several human-interpretable approaches to trust quantification which allow for a model's overall trustworthiness to be analyzed. These techniques are based on the notion of question-answer trust, in which a question $x$ (i.e., "is this solder joint defective?") is answered by both a model $M$ and an oracle $O$. The model's answer and confidence in its answer are compared to the oracle's answer to compute a question-answer trust score which reflects the trustworthiness of the model's answer.

To evaluate SolderNet's trustworthiness, we make use of two key trust metrics which summarize a model's overall question-answer trust:
\begin{enumerate}
    \item \textbf{Trust matrix}: a matrix of expected question-answer trusts for each possible model-oracle answer pair, and
    \item \textbf{NetTrustScore}: a scalar metric summarizing the overall trustworthiness of a deep neural network across all possible predictions.
\end{enumerate}

\subsection{Experimental Setup}

\subsubsection{Dataset}

To build and evaluate the proposed solder joint defect identification and explanation framework, we leverage a dataset comprising 2690 images of both through-hole and surface-mount joints acquired by Moog Inc. Images were acquired using a microscope camera and were labelled as either defective joints (1644 images) or non-defective joints (1046 images) by an experienced inspector. Notably, labelling was performed off-line without time constraints, thereby minimizing label noise in the resulting data. The dataset includes a variety of solder defects such as fractured solder joints, cold joints, burns, flux residue, poor wetting, and disturbed solder as illustrated in Figure~\ref{defects}. We used a 60\%/20\%/20\% split stratified by class to divide this dataset into training, validation, and test sets, respectively.

\subsubsection{Preprocessing}

Images are preprocessed by resizing them to 256$\times$256 pixels and mapping their original unsigned 8-bit integer pixel values to the range [0, 1] through division by 255. Additionally, the following data augmentations were used during training, each applied with a 50\% probability: random rotation in the range [0$^\circ$, 45$^\circ$], random horizontal and vertical flipping, random translation of $\pm$10\% in each direction, random brightness jitter of $\pm$20\%, and random contrast jitter of $\pm$20\%.

\subsubsection{Training}

All models examined in this work were pretrained on ImageNet-1k~\cite{imagenet}. Following~\cite{kumar2022finetuning}, each model's fully-connected layers were trained for 100 epochs followed by full-model fine-tuning for 1000 epochs. Binary cross-entropy loss and an AdamW optimizer~\cite{adamw} with $(\beta_1,\beta_2)=(0.9, 0.999)$ and weight decay of 1$\times10^{-4}$ were used in all experiments. We used a batch size of 128 and a learning rate of 1$\times10^{-3}$ to train the fully-connected layers, followed by full-network fine-tuning with a batch size of 128, an initial learning rate of 5$\times10^{-4}$, and cosine learning rate decay.

\subsubsection{Evaluation} To evaluate performance, we report accuracy, overkill rate (number of false-positives divided by number of samples), and escape rate (number of false-negatives divided by number of samples) on the the holdout test set. We chose these metrics due to the manufacturing context of this work; manufacturers care more about the absolute rates of false-positives and false-negatives rather than the proportional rates. Additionally, we report inference latency on an ARM Cortex-A72 processor, as well as NetTrustScore~\cite{trust3} to evaluate the trustworthiness of each model. Lastly, visual explanations and trust quantification plots were qualitatively evaluated.

\section{Results}

\subsection{Quantitative Results}

Quantitative performance metrics for each of the tested network architectures are shown in Table~\ref{table:quant}. All tested architectures achieve high performance, with Attend-NeXt~Large achieving the highest accuracy~(91.1\%) and lowest overkill rate~(5.0\%), MobileNetV2 achieving the lowest escape rate~(3.7\%), and Attend-NeXt~Small achieving the lowest latency~(0.275~s).

An intuitive interpretation of these metrics can be obtained by considering a scenario where 100 solder joints are to be inspected in a deployment setting. Considering the metrics of Attend-NeXt~Large, we see that it would take 43~s to inspect the set of joints and about 91/100 joints would be correctly classified. Of the remaining 9 misclassified joints, about 5 would be misclassified as defective and about 4 would be misclassified as non-defective. However, it is important to note that this interpretation is only meaningful if the proportions of defective and non-defective joints in the test data are representative of the true proportions in deployment. In this study, defective solder joints are over-represented in the test data, and as such we might expect higher overkill rates and reduced escape rates in practice.

\begin{figure}[t]
    \centering
    \includegraphics[width=0.7\columnwidth]{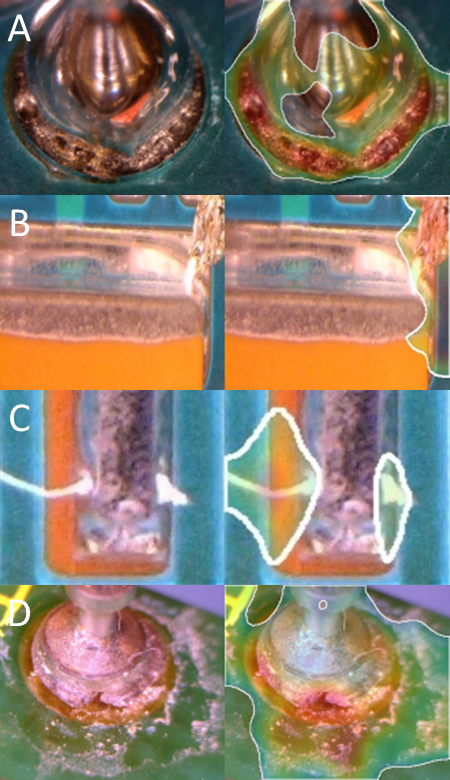}
    \caption{Images of solder joint defects (left) and corresponding visual explanations (right) from the dataset examined in this study: (A) poor wetting, (B) solder splash, (C) foreign object in solder, and (D) pad and board damage.}
    \label{fig:xai}
\end{figure}

\subsection{Explainability Results}

In this section we present and describe defect explanation and localization results obtained through the proposed explainability module. Figure~\ref{fig:xai} illustrates defective solder joints and the corresponding explanations of an Attend-NeXt Large model's predictions for these images. In each case, the XAI algorithm identified the critical factors that quantitatively drove the resulting decision (white outline) and the relative importance of aspects within each critical factor (semi-transparent regional heatmaps) indicating the critical factors used by the model to make its prediction. In all five cases, we observe that GSInquire identifies the key areas of interest driving the network's decision-making process and further localizes specific features via the regional heatmaps. These cases are described in more detail below.

\subsubsection{Case A} This image illustrates poor wetting, where a globule of solder which has not adequately joined with the solder pad can be seen. Additionally, the solder pad itself appears lumpy and discoloured which indicates possible residue or contamination. Examining the visual explanation, we see that the model correctly focuses on the non-wetted regions of the joint when making its prediction.

\subsubsection{Case B} In this example, the soldering process has introduced a solder splash which can be seen in the top-right corner of the image. Examining the visual explanation, we see that the model correctly identified this solder splash as the defect and effectively ignored the solder joint (which is well-formed) when making its decision.

\subsubsection{Case C} In this image, a foreign object (a piece of fibre) was inadvertently embedded in the solder when the joint was created. This foreign object is captured in its entirety by visual explanation, with the fibre itself being highlighted as the most important aspect in the explanation.

\subsubsection{Case D} Extensive damage to the pad and board are shown in this example, where a large piece has chipped off of the solder pad and extensive damage to the board's surface around the joint can be seen. While both of these aspects would constitute a defective joint, we see in the visual explanation that the model focuses on the damage to the solder pad while still including the surface damage.

\subsection{Second-order Explainability Results}

SOXAI visualizations are produced by placing the quantitative explanations produced by GSInquire across the wealth of data at hand at their corresponding 2-dimensional embedding locations following t-SNE. The resulting image illustrates groups of similar quantitative explanations which can more easily be examined for semantic groupings and common trends and patterns.

\begin{figure*}[t]
    \centering
    \includegraphics[width=0.7\textwidth]{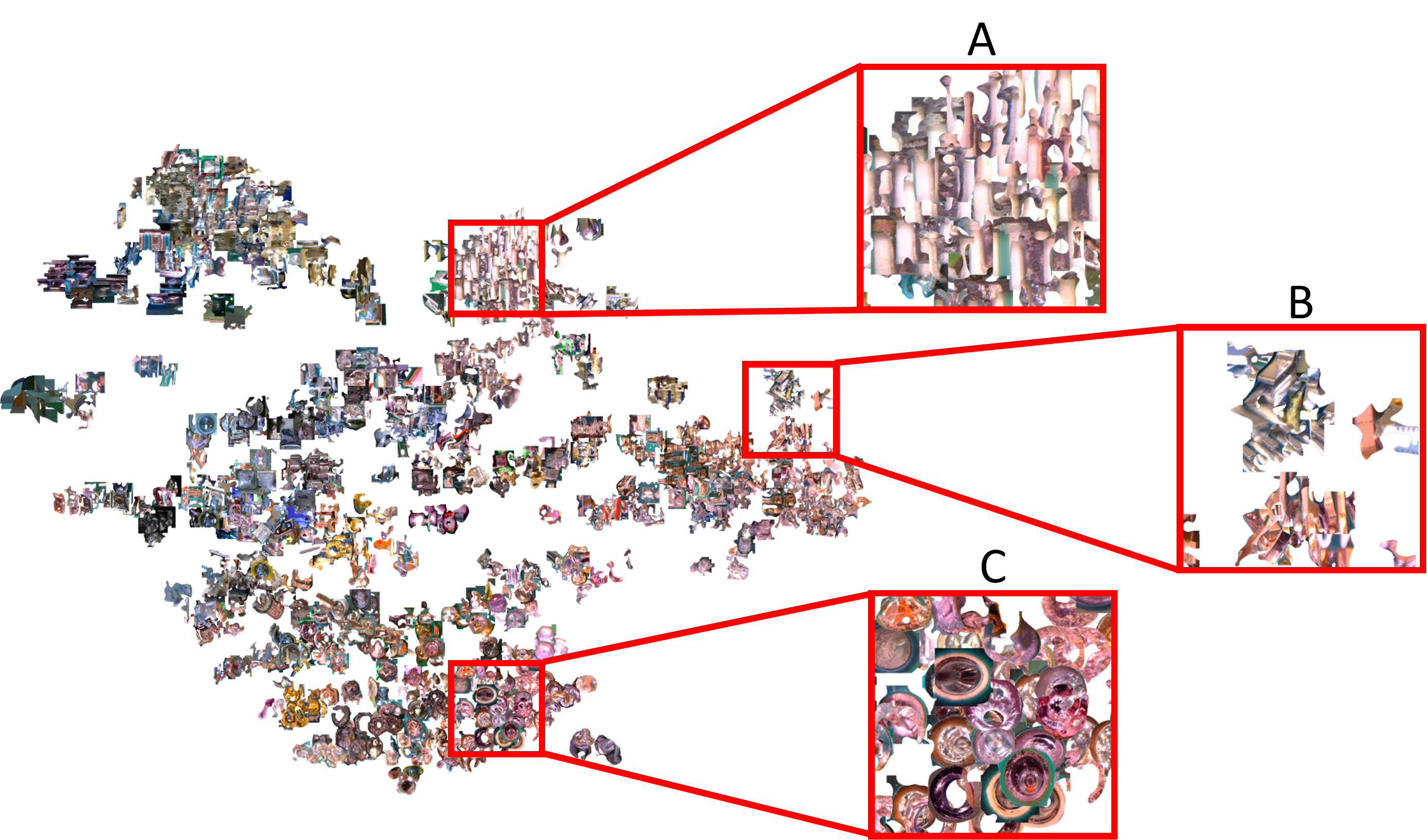}
    \caption{Second-order visual explainability illustrating various types of solder joint defects as viewed by the network: (A) side overhang, (B) lifted/unsoldered leads, and (C) wetting defects.}
    \label{fig:xai2nd}
\end{figure*}

Figure~\ref{fig:xai2nd} illustrates a SOXAI visualization for an Attend-NeXt Large model. As shown in the figure, SOXAI automatically groups explanations with similar characteristics, making it easier to find trends in the visual explanations. For example, in Figure~\ref{fig:xai2nd}~(A), we see that a homogeneous set of overhang defects has been tightly grouped. The relatively large size of this group indicates that this particular defect is well-recognized by the model but may be over-represented. In contrast, Figure~\ref{fig:xai2nd}~(B) shows a group of lifted leads which exhibit greater diversity but may be under-represented. In the neighbourhood of Figure~\ref{fig:xai2nd}~(C) we observe a large variety of through-hole defects, with (C) highlighting a group of wetting defects. The large size of the through-hole group paired with the intra-group variability indicates that through-hole joints and their defects are well-represented in the dataset and well-recognized by the model.

\begin{figure}[t]
    \centering
    \includegraphics[width=0.85\columnwidth]{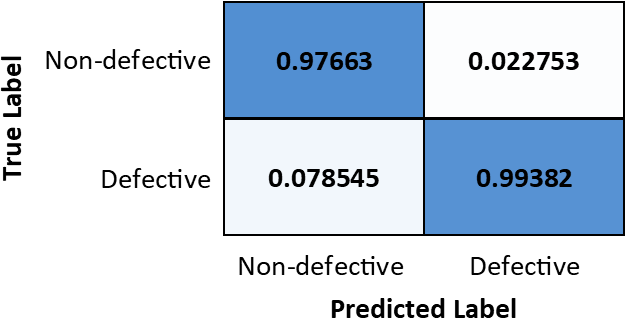}
    \caption{Trust matrix of Attend-NeXt Large.}
    \label{fig:trust}
\end{figure}

\subsection{Trust Analysis}

The NetTrustScores for the models examined in this work are shown in the rightmost column of Table~\ref{table:quant}. These scores give an overall measure of how trustworthy each model's predictions are. As shown, there is little disparity in trust amongst the models examined in this work, with Attend-NeXt Large achieving the highest score of 0.904 and Attend-NeXt Small having the lowest score of 0.864.

To explore trust in more detail, Figure~\ref{fig:trust} shows the trust matrix for Attend-NeXt Large. Notably, we show this matrix as an illustrative example since the trust matrices for the other architectures have a similar pattern. To interpret the trust matrix, consider that each entry indicates the expected question-answer trust for the given ground-truth/prediction pair. As such, higher values are better in all cells. Examining Figure~\ref{fig:trust}, we see that the diagonal entries (i.e., correct prediction scenarios) exhibit high trust. However, the off-diagonal entries (i.e., incorrect prediction scenarios) exhibit extremely low trust, indicating that the model is overconfident when it makes incorrect predictions. This is problematic in deployment scenarios, as it makes it more difficult to identify uncertain model predictions in order to flag them for manual review. To alleviate this problem, techniques such as label smoothing and mixup regularization~\cite{mixup} could be used to soften the image labels during training and encourage intermediate confidence scores.

\section{Discussion}

In this work, we described a design for a trustworthy, explainable solder joint inspection system for use in electronics manufacturing. While this system has yet to be fully implemented, we present important progress towards trustworthy, explainable solder joint inspection which forms the core of the proposed system. Moreover, we discuss practical considerations for building and evaluating such a system and show how trust quantification, quantitative explainability, and second-order explainability can be leveraged to analyze the trustworthiness of the system and identify biases or gaps in the data and model development process as well as provide insights during inspection.

The image data analyzed in this study varies considerably in terms of camera viewpoint, magnification, and resolution. In practice, a standardized imaging system would be required, however implementing an adequate imaging system is technically challenging due to the fact that different types of solder joints may need to be imaged at different angles, exposures, or resolutions in order to capture the majority of possible solder defects. In this study, we have assumed that such a system \textit{can} be designed, but the specific details of how to do so are left to future work.

When deploying a system such as the one described in this study, it is important to monitor and validate the system's performance in the field in order to identify and correct any issues that arise. No amount of offline testing can fully simulate a system's real-world performance, and so collecting prediction and performance data in the field is critical to evaluation. Additionally, false predictions observed in the field can be collected, curated, and used to fine-tune the system in order to reduce overkill and escape rates. Such continuous monitoring also helps to identify and mitigate drift in the system (for example, due to drift in camera calibration or other imaging parameters).

\section*{Acknowledgments}
We would like to thank Saeejith Nair for measuring the inference latencies of the architectures examined in this work.

\bibliography{references}

\begin{thebibliography}{25}
\providecommand{\natexlab}[1]{#1}

\bibitem[{Bhatt et~al.(2021)Bhatt, Malhan, Rajendran, Shah, Thakar, Yoon, and
  Gupta}]{autoinsp4}
Bhatt, P.~M.; Malhan, R.~K.; Rajendran, P.; Shah, B.~C.; Thakar, S.; Yoon,
  Y.~J.; and Gupta, S.~K. 2021.
\newblock {Image-Based Surface Defect Detection Using Deep Learning: A Review}.
\newblock \emph{Journal of Computing and Information Science in Engineering},
  21(4).

\bibitem[{Carratino et~al.(2020)Carratino, Cisse, Jenatton, and Vert}]{mixup}
Carratino, L.; Cisse, M.; Jenatton, R.; and Vert, J.-P. 2020.
\newblock On Mixup Regularization.
\newblock Technical report, arXiv.
\newblock 2006.06049.

\bibitem[{{Deng} et~al.(2009){Deng}, {Dong}, {Socher}, {Li}, {Kai Li}, and {Li
  Fei-Fei}}]{imagenet}
{Deng}, J.; {Dong}, W.; {Socher}, R.; {Li}, L.; {Kai Li}; and {Li Fei-Fei}.
  2009.
\newblock ImageNet: A large-scale hierarchical image database.
\newblock In \emph{2009 IEEE Conference on Computer Vision and Pattern
  Recognition}, 248--255.

\bibitem[{Erion et~al.(2021)Erion, Janizek, Sturmfels, Lundberg, and
  Lee}]{expgrad}
Erion, G.; Janizek, J.~D.; Sturmfels, P.; Lundberg, S.~M.; and Lee, S.-I. 2021.
\newblock Improving performance of deep learning models with axiomatic
  attribution priors and expected gradients.
\newblock \emph{Nature Machine Intelligence}, 3: 620–631.

\bibitem[{Howard et~al.(2019)Howard, Sandler, Chen, Wang, Chen, Tan, Chu,
  Vasudevan, Zhu, Pang, Adam, and Le}]{mobilenetv3}
Howard, A.; Sandler, M.; Chen, B.; Wang, W.; Chen, L.-C.; Tan, M.; Chu, G.;
  Vasudevan, V.; Zhu, Y.; Pang, R.; Adam, H.; and Le, Q. 2019.
\newblock Searching for MobileNetV3.
\newblock In \emph{2019 IEEE/CVF International Conference on Computer Vision
  (ICCV)}, 1314--1324.

\bibitem[{Hryniowski, Wong, and Wang(2021)}]{trust2}
Hryniowski, A.; Wong, A.; and Wang, X. 2021.
\newblock Where Does Trust Break Down? A Quantitative Trust Analysis of Deep
  Neural Networks via Trust Matrix and Conditional Trust Densities.
\newblock \emph{Journal of Computational Vision and Imaging Systems}, 6: 1--5.

\bibitem[{Kim et~al.(2021)Kim, Ko, Choi, and Kim}]{autoinsp1}
Kim, J.; Ko, J.; Choi, H.; and Kim, H. 2021.
\newblock {{P}rinted {C}ircuit {B}oard {D}efect {D}etection {U}sing {D}eep
  {L}earning via {A} {S}kip-{C}onnected {C}onvolutional {A}utoencoder}.
\newblock \emph{Sensors (Basel)}, 21(15).

\bibitem[{Klamklay and Bishu(1998)}]{inspection2}
Klamklay, J.; and Bishu, R.~R. 1998.
\newblock Visual Inspection of Circuit Boards: Effect of Gender, Age, Defect
  type, and Defect Proportion.
\newblock \emph{Proceedings of the Human Factors and Ergonomics Society Annual
  Meeting}, 42(16): 1161--1164.

\bibitem[{Kumar et~al.(2022)Kumar, Raghunathan, Jones, Ma, and
  Liang}]{kumar2022finetuning}
Kumar, A.; Raghunathan, A.; Jones, R.~M.; Ma, T.; and Liang, P. 2022.
\newblock Fine-Tuning can Distort Pretrained Features and Underperform
  Out-of-Distribution.
\newblock In \emph{International Conference on Learning Representations}.

\bibitem[{Lin et~al.(2019)Lin, Shafiee, Bochkarev, St.~Jules, Wang, and
  Wong}]{gsinquire}
Lin, Z.~Q.; Shafiee, M.~J.; Bochkarev, S.; St.~Jules, M.; Wang, X.~Y.; and
  Wong, A. 2019.
\newblock Do Explanations Reflect Decisions? A Machine-centric Strategy to
  Quantify the Performance of Explainability Algorithms.

\bibitem[{Liu et~al.(2022)Liu, Mao, Wu, Feichtenhofer, Darrell, and
  Xie}]{convnext}
Liu, Z.; Mao, H.; Wu, C.-Y.; Feichtenhofer, C.; Darrell, T.; and Xie, S. 2022.
\newblock A ConvNet for the 2020s.
\newblock \emph{Proceedings of the IEEE/CVF Conference on Computer Vision and
  Pattern Recognition (CVPR)}.

\bibitem[{Loshchilov and Hutter(2019)}]{adamw}
Loshchilov, I.; and Hutter, F. 2019.
\newblock Decoupled Weight Decay Regularization.
\newblock In \emph{International Conference on Learning Representations}.

\bibitem[{Lundberg and Lee(2017)}]{shap}
Lundberg, S.; and Lee, S.-I. 2017.
\newblock A Unified Approach to Interpreting Model Predictions.
\newblock In \emph{Proceedings of the 31st International Conference on Neural
  Information Processing Systems}, 768–4777.

\bibitem[{Ma et~al.(2018)Ma, Zhang, Zheng, and Sun}]{shufflenetv2}
Ma, N.; Zhang, X.; Zheng, H.-T.; and Sun, J. 2018.
\newblock {ShuffleNet V2}: Practical Guidelines for Efficient {CNN}
  Architecture Design.
\newblock In Ferrari, V.; Hebert, M.; Sminchisescu, C.; and Weiss, Y., eds.,
  \emph{Computer Vision -- ECCV 2018}, 122--138. Cham: Springer International
  Publishing.
\newblock ISBN 978-3-030-01264-9.

\bibitem[{Ribeiro, Singh, and Guestrin(2016)}]{lime}
Ribeiro, M.; Singh, S.; and Guestrin, C. 2016.
\newblock {``}Why Should {I} Trust You?{''}: {E}xplaining the Predictions of
  Any Classifier.
\newblock In \emph{Proceedings of the 2016 Conference of the North {A}merican
  Chapter of the Association for Computational Linguistics: Demonstrations},
  97--101. San Diego, California: Association for Computational Linguistics.

\bibitem[{Sandler et~al.(2018)Sandler, Howard, Zhu, Zhmoginov, and
  Chen}]{mobilenetv2}
Sandler, M.; Howard, A.; Zhu, M.; Zhmoginov, A.; and Chen, L.-C. 2018.
\newblock {MobileNetV2}: {I}nverted Residuals and Linear Bottlenecks.
\newblock In \emph{2018 IEEE/CVF Conference on Computer Vision and Pattern
  Recognition}, 4510--4520.

\bibitem[{See et~al.(2017)See, Drury, Speed, Williams, and
  Khalandi}]{inspection1}
See, J.~E.; Drury, C.~G.; Speed, A.; Williams, A.; and Khalandi, N. 2017.
\newblock The Role of Visual Inspection in the 21st Century.
\newblock \emph{Proceedings of the Human Factors and Ergonomics Society Annual
  Meeting}, 61(1): 262--266.

\bibitem[{{Selvaraju} et~al.(2017){Selvaraju}, {Cogswell}, {Das}, {Vedantam},
  {Parikh}, and {Batra}}]{gradcam}
{Selvaraju}, R.~R.; {Cogswell}, M.; {Das}, A.; {Vedantam}, R.; {Parikh}, D.;
  and {Batra}, D. 2017.
\newblock Grad-CAM: Visual Explanations from Deep Networks via Gradient-Based
  Localization.
\newblock In \emph{2017 IEEE International Conference on Computer Vision
  (ICCV)}, 618--626.

\bibitem[{van~der Maaten and Hinton(2008)}]{tsne}
van~der Maaten, L.; and Hinton, G. 2008.
\newblock Visualizing Data using t-SNE.
\newblock \emph{Journal of Machine Learning Research}, 9(86): 2579--2605.

\bibitem[{Westphal and Seitz(2021)}]{autoinsp3}
Westphal, E.; and Seitz, H. 2021.
\newblock A machine learning method for defect detection and visualization in
  selective laser sintering based on convolutional neural networks.
\newblock \emph{Additive Manufacturing}, 41: 101965.

\bibitem[{Wong, Hryniowski, and Wang(2020)}]{trust1}
Wong, A.; Hryniowski, A.; and Wang, X.~Y. 2020.
\newblock Insights into Fairness through Trust: Multi-scale Trust
  Quantification for Financial Deep Learning.

\bibitem[{Wong et~al.(2022)Wong, Shafiee, Abbasi, Nair, and
  Famouri}]{attendnext}
Wong, A.; Shafiee, M.~J.; Abbasi, S.; Nair, S.; and Famouri, M. 2022.
\newblock Faster Attention Is What You Need: {A} Fast Self-Attention Neural
  Network Backbone Architecture for the Edge via Double-Condensing Attention
  Condensers.

\bibitem[{Wong, Wang, and Hryniowski(2020)}]{trust3}
Wong, A.; Wang, X.~Y.; and Hryniowski, A. 2020.
\newblock How Much Can We Really Trust You? Towards Simple, Interpretable Trust
  Quantification Metrics for Deep Neural Networks.

\bibitem[{Yang et~al.(2020)Yang, Li, Wang, Dong, Wang, and Tang}]{autoinsp5}
Yang, J.; Li, S.; Wang, Z.; Dong, H.; Wang, J.; and Tang, S. 2020.
\newblock Using Deep Learning to Detect Defects in Manufacturing: A
  Comprehensive Survey and Current Challenges.
\newblock \emph{Materials}, 13(24).

\bibitem[{Zhang et~al.(2022)Zhang, Zhang, Gamanayake, Yuen, Geng, Jayasekara,
  wei Woo, Low, Liu, and Guan}]{autoinsp2}
Zhang, Q.; Zhang, M.; Gamanayake, C.; Yuen, C.; Geng, Z.; Jayasekara, H.; wei
  Woo, C.; Low, J.; Liu, X.; and Guan, Y.~L. 2022.
\newblock {Deep learning based solder joint defect detection on industrial
  printed circuit board X-ray images}.
\newblock \emph{Complex \& Intelligent Systems}, 8: 1525--1537.

\end{thebibliography}

\end{document}